\documentclass[letterpaper]{article}
\PassOptionsToPackage{table}{xcolor}
\usepackage{aaai2026}
\nocopyright 
\usepackage{times}
\usepackage{helvet}
\usepackage{courier}
\usepackage[hyphens]{url}
\usepackage{graphicx}
\usepackage{natbib}
\usepackage{caption}
\usepackage{booktabs}
\usepackage{multirow}
\usepackage{amsmath}
\usepackage{amssymb}
\usepackage{pifont}
\usepackage{xcolor}
\newcommand{\smat}{S_{\mathrm{mat}}}
\newcommand{\spres}{S_{\mathrm{pres}}}
\newcommand{\sstr}{S_{\mathrm{struct}}}
\newcommand{\sharm}{S_{\mathrm{harm}}}

\title{MatReplace: A Reference-Free, Conditioning-Aligned Benchmark for\\Material Replacement in Interior Scenes}
\author{
    Mingzhe~Du\textsuperscript{\rm 1},
    Thong~Thanh~Nguyen\thanks{Corresponding author.}\textsuperscript{\rm 1},
    Nguyen~Tran~Cong~Duy\textsuperscript{\rm 2},
    \mbox{See-Kiong}~Ng\textsuperscript{\rm 1},
    Luu~Anh~Tuan\textsuperscript{\rm 2,3}
}
\affiliations{
    \textsuperscript{\rm 1}National University of Singapore \quad
    \textsuperscript{\rm 2}Nanyang Technological University \quad
    \textsuperscript{\rm 3}VinUniversity \\
    \{mingzhe, thong.nguyen, seekiong\}@nus.edu.sg, \{nguyentr003, anhtuan.luu\}@ntu.edu.sg
}

\begin{document}
\maketitle

\begin{abstract}
Generative image editors are entering interior-design workflows, where the routine request is \emph{material replacement}: repaint one marked surface with a new material while the object's geometry, the surrounding scene, and the illumination stay fixed.
The operation is precisely specified and commercially routine, yet no public benchmark isolates it, and scoring it is subtle. The task is one-to-many, so distance to a stored reference penalizes valid diversity; a single-generator source rewards imitating that generator's style; and editors given different guidance cannot share one leaderboard fairly.
We introduce \textbf{MatReplace}, a reference-free benchmark that scores each edit on four verifiable dimensions~(\emph{local material correctness, global lighting harmony, outside preservation, inside structure}), across three tracks that vary one conditioning signal at a time: (A)~the instruction alone, (B)~the instruction plus a region mask, and (C)~a reference image of the material in place of the instruction.
Track A shows that strong closed-source models already render the requested material at exemplar quality, clearing the exemplar anchor under our primary aggregate. Track B shows that masks help only models that preserve the scene poorly and can consume them: their task-paired, single-seed value spans $+0.137$ to $-0.090$ across aligned families. Track C degrades every family under both aggregates, by $-0.031$ to $-0.508$, at worst repainting the reference itself and dropping below the score of returning the input untouched.
Rendering a \emph{named material} is thus largely solved for the strongest closed editors on this distribution, while grounding a material shown as \emph{pixels} is not. Expert human ratings reproduce our ranking~(Kendall $\tau=0.68$) and track our aggregates more closely than GT-referenced or CLIP-based baselines.
\end{abstract}

\section{Introduction}
Interior-design tools increasingly use generative editors to answer a concrete client question: \emph{what would this room look like if the headboard were navy suede, or the floor herringbone oak?} The underlying operation, \textbf{material replacement}, is clearly specified.
The material of one marked surface must change to a stated target while the object's geometry, the rest of the scene, and the scene's illumination remain fixed.
Despite its practical importance and clear definition, no existing benchmark isolates this task. Instruction-editing suites~\cite{zhang2023magicbrush,sheynin2024emu,wang2023imagen} fold it into broad edit categories, and their metrics score edits against a single stored reference image per task.

\begin{figure*}[t]
    \centering
    \includegraphics[width=0.98\textwidth]{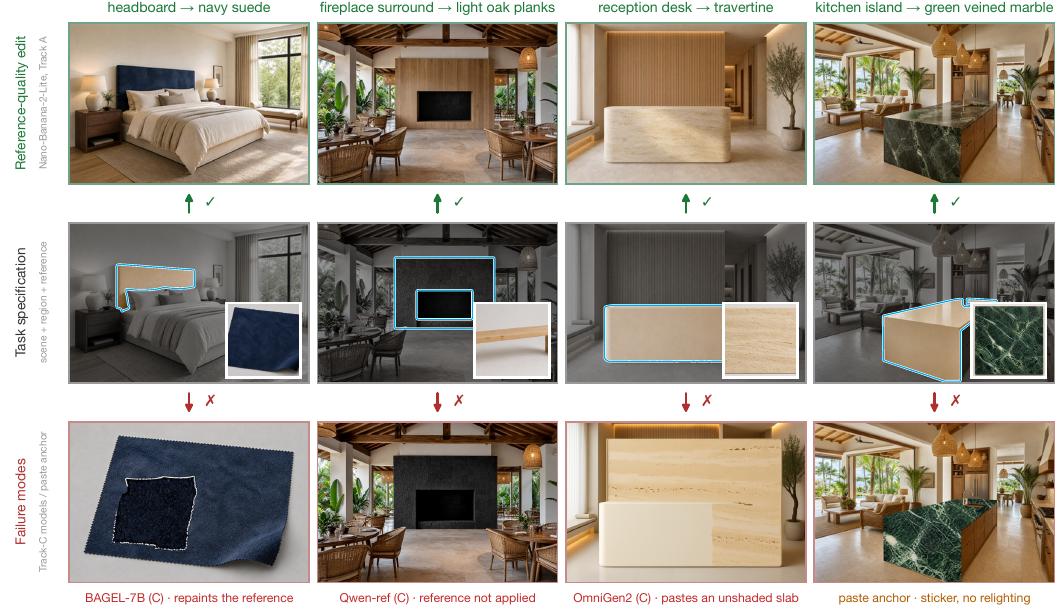}
    \caption{
        \textbf{MatReplace at a glance.} Each column is one task, read from the middle out.
        \textbf{Middle:} the target region in color, outlined over a dimmed grayscale surround (mask $m$), reference $r$ inset.
        \textbf{Up} ({\color[HTML]{1B7837}\checkmark}): a reference-quality edit (Nano-Banana-2-Lite, Track~A). Material changes while geometry, scene, and illumination survive.
        \textbf{Down} ({\color[HTML]{B03030}\ding{55}}): one failure per mode. 
        BAGEL-7B (C) repaints the \emph{reference} and discards the scene (column 1); 
        Qwen-Image-Edit (C) fails to apply the reference and leaves the slate untouched (column 2); 
        OmniGen2 (C) pastes a flat, unshaded slab (column 3); 
        \emph{Paste anchor} tiles the reference without relighting (column 4). 
        Each failure is caught by the dimension whose contract it violates (Sec.~\ref{sec:protocol}).
    }
    \label{fig:showcase}
\end{figure*}

Scoring against stored reference images is doubly problematic for this task. First, the task is one-to-many. Many navy-suede headboards are equally correct, and penalizing distance to one stored example punishes valid diversity. Second, when all stored reference images come from a single generator, any such metric rewards matching that generator's rendering style and biases the leaderboard toward that data source. MatReplace therefore treats the stored output as \emph{one reference solution} that supports qualitative comparison and calibration, never the scoring target. Evaluation instead decomposes the task definition into four independently verifiable properties, each checked only against the provided inputs: whether the masked region became the target material, whether the outside stayed unchanged, whether the inside geometry survived, and whether the new surface matches the scene illumination.

The second design choice concerns \emph{conditioning}, the form of guidance a system receives. Some deployed editors receive only a text instruction, some additionally receive a region mask, and others receive a mask together with a reference image of the target material. Benchmarks mixing these on one leaderboard conflate model quality with input signals. MatReplace defines Track~A (instruction), Track~B (instruction$+$mask), and Track~C (mask$+$reference), aligning the same model family across tracks wherever the architecture permits. Holding the model fixed and varying only conditioning measures what a mask or reference is worth; the answer is neither uniform nor always positive. Our contributions are:
\begin{itemize}
    \item \textbf{MatReplace}, a material-replacement benchmark: a frozen split over several material categories and surface types, with taxonomy labels and released inputs.
    \item A \textbf{reference-free protocol} of four independently verifiable dimensions, validated by constructed submissions that incur penalties only on their target.
    \item A \textbf{three-track leaderboard} defined by conditioning, with families aligned across tracks. Masks help only families weak at preservation and able to consume them; reference images degrade all models, in three failure modes the dimensions distinguish.
\end{itemize}

\section{Related Work}
\paragraph{Instruction-based editing and its benchmarks.}
End-to-end instruction editing has produced many manual and large-scale benchmarks~\cite{brooks2023instructpix2pix,zhang2023magicbrush,sheynin2024emu,wang2023imagen,kawar2023imagic,ku2024imagenhub}, all scoring primarily by GT-referenced distance plus CLIP text alignment~\cite{hessel2021clipscore} and inheriting both biases above. Recent flow-matching and unified any-to-any editors~\cite{labs2025flux,wu2025qwen,wu2026omnigen2,deng2025emerging} motivate our aligned design: one checkpoint enters all three tracks.

\paragraph{Reference-guided appearance transfer.}
Exemplar-based inpainting and object customization insert a reference's content into a scene~\cite{yang2023paint,chen2024anydoor,ruiz2023dreambooth}. Track~C is the material-level analogue: the reference specifies \emph{appearance} (color, texture, micro-structure) to be re-rendered under the scene's geometry and illumination, not pasted; harmony and structure separate the two.

\paragraph{Material editing and design-domain data.}
A parallel line edits materials without benchmarking the task: parametric control~\cite{sharma2024alchemist}, zero-shot single-exemplar transfer~\cite{cheng2024zest}, light-aware transfer~\cite{lopes2025matswap}, and material-map extraction~\cite{lopes2024material}, each on its own small set. No shared suite compares them with general editors, the gap MatReplace fills; despite interiors synthesized at scale from 3D-scene repositories~\cite{fu20213d,roberts2021hypersim}, no leaderboard isolates an interior editing operation.

\paragraph{Material understanding and intrinsics.}
Material recognition in situ is a classic vision problem~\cite{bell2013opensurfaces,bell2015material}, and the albedo--shading decomposition underlying our harmony dimension goes back to intrinsic-image analysis~\cite{grosse2009ground}. MatReplace connects this line to generative editing: material correctness via zero-shot SigLIP2~\cite{tschannen2025siglip} recognition and DINOv2~\cite{oquab2023dinov2} appearance similarity, and harmony via shading invariance.

\paragraph{Reference-free evaluation.}
CLIPScore~\cite{hessel2021clipscore} showed reference-free scoring for captioning, a single-obligation task. For editing the right unit is a \emph{decomposition into the task's contract}: each dimension ties to a checkable clause, built from established components (LPIPS~\cite{zhang2018unreasonable}, SSIM~\cite{wang2004image}, Depth Anything V2~\cite{yang2024depth}) and validated by an anchor experiment. Unlike a learned judge, every dimension is auditable and deterministic with fixed, released constants.

\section{The MatReplace Benchmark}
\paragraph{Task definition.}
A task is a tuple $(x, m, r, T, c)$: scene $x$, binary mask $m$ on one surface, reference $r$ of the target material, instruction $T$, and material-category label $c$ (10-way). A submission is an edited image $\hat{y}$ at $x$'s resolution. The contract: inside $m$ the material becomes $c$ (matching $r$ where provided); outside $m$ the scene is unchanged; the geometry inside is unchanged and the new surface shaded by the scene's illumination.

\paragraph{Why interior scenes.}
Interior visualization is where material replacement is deployed: design tools iterate over finishes on fixed geometry, so the edit must be surgical. It is also a stress test, with planar surfaces beside upholstery and drapery, strong mixed lighting, and material--object bindings dense enough that text struggles to pin them down.

\begin{figure}[t]
    \centering
    \includegraphics[width=\columnwidth]{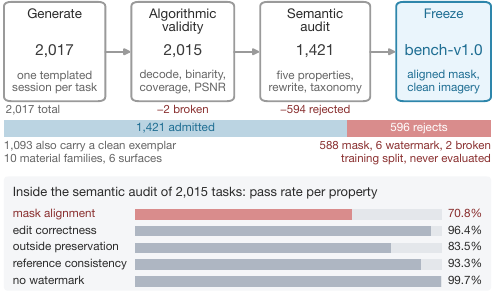}
    \caption{
        MatReplace Benchmark Construction. 
        One templated generation session per candidate task, an algorithmic validity pass, and a semantic audit by a vision--language agent reduce 2{,}017 candidate sessions to the frozen 1{,}421 tasks. The middle bar is drawn to scale, so the admitted and rejected shares are directly comparable and the accounting closes. The 596 rejects ship as a disjoint \emph{rejected pool}, released for auditing and never used for evaluation. 
    }
    \label{fig:pipeline}
\end{figure}

\paragraph{Sources and curation.}
Figure~\ref{fig:pipeline} summarizes construction. Each of 2{,}017 candidate tasks is one templated session: a GPT-5.6-Sol agent drives one disclosed generator, GPT-Image-2 (itself evaluated), producing the scene, the mask \emph{as an edit on that scene} (binarized, coverage-checked, regenerated on failure), the reference, an exemplar output, and a canonical instruction. An \emph{algorithmic} pass checks decodability, resolution agreement, mask binarity and in-bounds coverage, and flags exemplars that redraw the scene (low outside-mask PSNR); 2{,}015 survive. A \emph{semantic} pass then audits five properties with a GPT-5.6-Sol agent: mask--object alignment ($70.8\%$ pass), edit correctness ($96.4\%$), outside preservation ($83.5\%$), reference--edit consistency ($93.3\%$), and watermark absence ($99.7\%$), rewriting each instruction to match the realized images. Admission requires an aligned mask and clean imagery; defects in the stored \emph{exemplar} do not disqualify a task, since scoring never references it. Mask alignment is the dominant yield limiter; the 596 rejects ship as a disjoint \emph{rejected pool}, released for auditing rather than as supervision.

\paragraph{What the rejects look like.}
The rejects catalogue how a strong generator fails at its own task: masks spill or miss (588, dominant); edits leak outside the mask (332); inside geometry is re-sculpted (73); exemplars mismatch their reference (135) or redraw the scene (153), exactly the behaviors the four dimensions target (Fig.~\ref{fig:failmodes}).

\paragraph{Construction cost.}
At public GPT-Image-2 list prices (July 2026), a four-call session costs \$0.37--\$1.48, or \$0.7k--\$3.0k across 2{,}017 sessions; absorbing curation yield and the audit, an admitted task costs \$0.57--\$2.14, and the fully audited benchmark prices below a single small user study.

\paragraph{Statistics.}
MatReplace has 1{,}421 task specifications, spanning 10 material categories (fabric 187, wood 181, marble 172, stone 170, metal 160, \dots) and 6 surface types (furniture 825, wall 156, countertop 144, fireplace 125, \dots), with mask-coverage buckets retained for stratified analysis.

\paragraph{Three conditioning tracks.}
Tracks differ \emph{only} in inputs: \textbf{Track A} $x{+}T$, \textbf{Track B} $x{+}m{+}T$, and \textbf{Track C} $x{+}m{+}r$, the material named only by the image. A family supporting several modes enters several tracks from one checkpoint.

\paragraph{Scope and non-goals.}
MatReplace isolates \emph{one} edit type; insertion, removal, geometry edits, multi-surface restyling, and global style transfer are out of scope. A single-surface edit is the smallest unit for which preservation, structure, and harmony are simultaneously well defined, whereas composite edits reintroduce ambiguities needing a stored answer.

\section{Reference-Free Evaluation Protocol}
\label{sec:protocol}
\begin{figure}[t]
    \centering
    \includegraphics[width=\columnwidth]{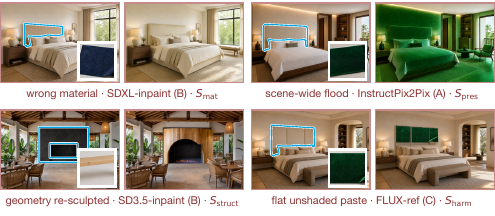}
    \caption{One failure per dimension: each pair shows the input with the target region outlined, then the model output. SDXL-inpaint paints bare wall instead of navy suede ($\smat$). InstructPix2Pix floods the whole room with the target emerald ($\spres$). SD3.5-inpaint re-sculpts the rectangular fireplace into an arched portal ($\sstr$). FLUX-Kontext-ref pastes the reference as flat, unlit patches ($\sharm$).}
    \label{fig:failmodes}
\end{figure}

Given $(x, m, r, c)$ and a submission $\hat{y}$, we compute four scores in $[0,1]$; \texttt{overall} is their unweighted mean. \emph{Reference-free} means free of any reference \emph{output}: every score is a function of the task inputs alone, never of a stored solution. The reference $r$ is itself an input; its Track-A consequence is discussed under the design rationale. Let $\tilde{m}$ denote $m$ dilated by a $7$-px boundary band (excluded from outside comparisons), $\bar{m}=1-\tilde{m}$ its complement, and $\mathrm{crop}(\cdot, m)$ the masked crop. Figure~\ref{fig:failmodes} shows outputs violating exactly that clause.

\paragraph{Material correctness ($\smat$).}
SigLIP2 classification $\hat{c}$ of $\mathrm{crop}(\hat{y},m)$ over the material vocabulary, scored at the \emph{family} level (pairs that vision encoders legitimately confuse, such as marble and granite, are grouped), averaged with a DINOv2 ($f_{\mathrm{D}}$) cosine similarity between the crop and $r$:
\begin{equation}
\begin{split}
\smat = \tfrac{1}{2}\Big[&\mathbf{1}\{\mathrm{fam}(\hat{c}) = \mathrm{fam}(c)\}\\
 &+ \cos\!\big(f_{\mathrm{D}}(\mathrm{crop}(\hat{y},m)),\, f_{\mathrm{D}}(r)\big)\Big].
\end{split}
\end{equation}

\paragraph{Outside preservation ($\spres$).}
LPIPS between $x$ and $\hat{y}$ restricted to the outside region:
\begin{equation}
\spres = \max\big(0,\, 1-\mathrm{LPIPS}(x \odot \bar{m},\, \hat{y} \odot \bar{m})\big),
\end{equation}
with PSNR/SSIM reported as diagnostics. $\spres$ is invariant to the resize-back-to-input-resolution step applied to all submissions (round-trip verified).

\paragraph{Inside structure ($\sstr$).}
Monocular depth $d(\cdot)$ for $x$ and $\hat{y}$, aligned by scale-and-shift on the \emph{outside} region, then the inside absolute-relative error over $m$:
\begin{equation}
\sstr = \exp\!\big(-\mathrm{AbsRel}_{m}\big(d(x),\, a\,d(\hat{y})+b\big)/\tau\big),
\end{equation}
with temperature $\tau$ fixed across all models (all constants in the released harness).

\paragraph{Lighting harmony ($\sharm$).}
A material swap rescales albedo but must preserve the spatial \emph{pattern} of illumination. We correlate the low-pass luminance fields inside the mask,
\begin{equation}
\mathrm{NCC} = \mathrm{corr}\big(\ell_{\sigma}(x)\!\mid_m,\;
\ell_{\sigma}(\hat{y})\!\mid_m\big),
\end{equation}
which is invariant to the albedo scale, and average (after mapping to $[0,1]$) with a \emph{seam} term that penalizes gradient energy appearing just outside the mask boundary (halos, glow) relative to $x$.

\paragraph{Design rationale.}
Each dimension makes one trade. 
Material is scored at the family level because fine-grained labels are not reliably separable by current encoders even on clean crops, so strict scoring would grade the \emph{scorer}, not the model. The DINOv2 term anchors appearance to $r$ even on Track~A, where the model never sees $r$: two valid navy suedes differ in similarity to $r$, readmitting a bounded one-to-many penalty (at most half of $\smat$, one eighth of \texttt{overall}). We keep it because the rewritten instruction describes $r$'s appearance and removing it costs agreement with the human ranking (Table~\ref{tab:baselines}). Preservation excludes the boundary band because mask-edge pixels are ambiguous under resampling. Structure aligns depth on the \emph{outside} region, isolating inside edits from global depth-scale drift; monocular depth is not perfectly texture-invariant, so a correct swap costs a little structure (the exemplar loses $0.15$), against which $\tau$ is set. Harmony correlates \emph{low-pass} luminance, since an albedo change rescales brightness but must not restructure illumination, with a seam term watching the ring outside the mask where halos concentrate. Finally, \texttt{overall} is unweighted, so from the released per-task, per-dimension scores we read results dimension-wise.

\paragraph{Uncertainty and significance.}
Every mean carries a 95\% percentile-bootstrap interval over tasks ($B{=}10{,}000$, fixed seed). 
Since all submissions share the same tasks, deltas and near-ties are tested with two-sided Wilcoxon signed-rank on per-task overall scores, Holm-corrected across all 14 comparisons.
Each entry is a single fixed-seed run, so deltas below $0.02$ are significant over task sampling but within plausible seed variation, and we flag them ($\ddagger$) in Table~\ref{tab:cross}.

\paragraph{Anchor validation.}
Because the metrics are new, we validate them with three constructed submissions on all 1{,}421 tasks (Table~\ref{tab:anchors}): \emph{identity} (submit $x$) saturates preservation/structure/harmony and fails only material ($1.000/0.999/1.000$, material $0.242$); \emph{paste} (tile $r$, no re-rendering) takes the highest raw material score ($0.673$) but is caught by structure ($0.548$) and harmony ($0.697$), the ``sticker'' artifact; \emph{gt} (the exemplar) scores well on all four ($0.829$) as a soft ceiling; it averages every stored exemplar, including the 328 audit-flagged ones, and moves only to $0.833$ on the clean subset. The anchors span the observed range. The exemplar also calibrates conservatism: a correct change costs about $0.15$ structure and $0.12$ harmony (depth and low-pass luminance are not fully texture-invariant), compressing raw \texttt{overall} between floor $0.810$ and ceiling $0.829$.

\begin{table}[t]
    \centering
    \small
    \setlength{\tabcolsep}{3pt}
    \begin{tabular*}{\linewidth}{@{\extracolsep{\fill}}lccccc}
    \toprule
    anchor & mat & pres & struct & harm & overall\\
    \midrule
    gt (exemplar) & 0.622 & 0.964 & 0.853 & 0.877 & 0.829\,{\scriptsize$\pm$.004}\\
    identity (input)   & \textbf{0.242} & 1.000 & 0.999 & 1.000 & 0.810\,{\scriptsize$\pm$.003}\\
    paste (no relight) & 0.673 & 0.951 & \textbf{0.548} & \textbf{0.697} & 0.717\,{\scriptsize$\pm$.005}\\
    \bottomrule
    \end{tabular*}
    \caption{Anchor validation on all 1{,}421 tasks. Bold marks the dimension each anchor is constructed to violate, and no anchor is penalized elsewhere.}
    \label{tab:anchors}
    \vspace{-5pt}
\end{table}

\paragraph{Metric robustness.}
Raw \texttt{overall} compresses near identity, so we compare it against two alternatives (Table~\ref{tab:robust}). \emph{(i)}~Scoring $\smat$ by family accuracy alone (no DINOv2 term) barely moves the ranking ($\tau=0.91$) and lifts the exemplar above \emph{paste} on material. \emph{(ii)}~A gate $\mathrm{overall}_G=\hat{S}_{\mathrm{mat}}\cdot(\spres\,\sstr\,\sharm)^{1/3}$, with $\hat{S}_{\mathrm{mat}}=\mathrm{clip}\big((\smat-0.242)/0.380,\,0,\,1\big)$ the per-task material score rescaled by the global identity and exemplar anchor means, sends the do-nothing anchor to $0.236$ and lifts every entry above it except OmniGen2-ref ($0.204$) and BAGEL-7B ($0.062$), both Track C, at a mild re-ranking ($\tau=0.77$).
Both agree with the raw ranking on the top of the board and on the Track-C collapse: the compression is a property of the equal-weight mean among strong models, not of the ranking. We therefore report $\mathrm{overall}_G$ as the \emph{primary} aggregate and raw \texttt{overall} beside it. The gate is also the view closest to human judgment: it agrees best with the Bradley--Terry strengths of Sec.~\ref{sec:human} ($\tau_{\mathrm{H}}=0.72$ vs.\ $0.68$ raw and $0.64$ family-only; Table~\ref{tab:baselines}) and it ranks \emph{paste} above \emph{identity} as the raters do (6--3), unlike raw \texttt{overall}. All three aggregates were pre-specified, and the gate's constants are anchor means rather than parameters fitted to the judgments, so the agreement is a check, not a fit.

\begin{table}[t]
\centering
\small
\setlength{\tabcolsep}{4pt}
\begin{tabular*}{\linewidth}{@{\extracolsep{\fill}}lccccc}
\toprule
aggregate & identity & paste & gt & $>$id & $\tau_{\mathrm{raw}}$\\
\midrule
raw \texttt{overall}          & 0.810 & 0.717 & 0.829 & 4/23  & 1.00\\
$\smat$ family-only           & 0.813 & 0.749 & 0.882 & 9/23  & 0.91\\
gate $\mathrm{overall}_G$     & \textbf{0.236} & 0.552 & 0.729 & \textbf{21/23} & 0.77\\
\bottomrule
\end{tabular*}
\caption{
    Robustness of the aggregate. Columns give the three anchors, the number of the $23$ entries scoring above the \emph{identity} floor, and Kendall $\tau$ against the raw ranking. Correlations with the human ranking appear in Table~\ref{tab:baselines}.
}
\label{tab:robust}
\end{table}

\section{Experiments}
\paragraph{Evaluated systems.}
We evaluate 11 systems in 23 track entries, and six families complete all three tracks from one checkpoint or pipeline family. \emph{Closed} (provider APIs) include GPT-Image-2 (A/B/C, the disclosed data generator) and Nano-Banana-2-Lite (A/B/C). \emph{Open, cross-track} include FLUX.1-Kontext~\cite{labs2025flux} (A, inpainting for B, reference mode for C), Qwen-Image-Edit~\cite{wu2025qwen}, OmniGen2~\cite{wu2026omnigen2}, and BAGEL-7B~\cite{deng2025emerging} (all natively multi-image, A/B/C). \emph{Open, single-track} include InstructPix2Pix~\cite{brooks2023instructpix2pix} (A), FLUX.1-Fill, SDXL-inpainting~\cite{podell2024sdxl}, and SD3.5-inpainting~\cite{esser2024scaling} (B), and Paint-by-Example~\cite{yang2023paint} (C).

\paragraph{Protocol.}
All open models run at native or recommended resolution with default guidance and 50 denoising steps, seed fixed. Outputs are resized to the input resolution before scoring. Track prompts are templated identically across models: Track~B prepends the mask semantics, and Track~C names no material in text. Templating is minimal and untuned, since per-model prompt engineering would reintroduce the input-privilege confound the tracks remove. Track-C reference injection follows each model's native interface: the multi-image editors (Qwen-Image-Edit, OmniGen2, BAGEL) take an ordered list $[x, m, r]$; FLUX-Kontext uses its reference port ($\texttt{image\_reference}{=}r$) beside $\texttt{image}/\texttt{mask\_image}$; Paint-by-Example its exemplar port ($\texttt{example\_image}{=}r$), no text. The list-based prompts name the roles explicitly (``\emph{Image 3 shows a target material}''), so role underspecification is not the immediate cause of BAGEL's failure below.

\begin{table*}[t]
    \centering
    \footnotesize
    \renewcommand{\arraystretch}{0.93}
    \begin{tabular*}{\textwidth}{@{\extracolsep{\fill}}lccccccc}
        \toprule
        model & family acc. & material & preservation & structure & harmony & raw & gate\\
        \midrule
        \rowcolor{gray!18}\multicolumn{8}{l}{\textbf{Track A: instruction (input $+$ text)}}\\
        Nano-Banana-2-Lite$^\dagger$ & \textbf{0.899} & \textbf{0.640\,{\scriptsize$\pm$.011}} & \underline{0.953\,{\scriptsize$\pm$.001}} & \underline{0.820\,{\scriptsize$\pm$.007}} & 0.857\,{\scriptsize$\pm$.007} & \underline{0.817\,{\scriptsize$\pm$.004}} & \textbf{0.760}\\
        GPT-Image-2$^\dagger$ & 0.852 & \underline{0.616\,{\scriptsize$\pm$.012}} & \textbf{0.962\,{\scriptsize$\pm$.001}} & \textbf{0.853\,{\scriptsize$\pm$.006}} & \underline{0.878\,{\scriptsize$\pm$.007}} & \textbf{0.827\,{\scriptsize$\pm$.004}} & \underline{0.739}\\
        Qwen-Image-Edit & \underline{0.887} & 0.615\,{\scriptsize$\pm$.011} & 0.899\,{\scriptsize$\pm$.002} & 0.710\,{\scriptsize$\pm$.010} & 0.817\,{\scriptsize$\pm$.008} & 0.760\,{\scriptsize$\pm$.005} & 0.677\\
        BAGEL-7B & 0.797 & 0.533\,{\scriptsize$\pm$.012} & 0.917\,{\scriptsize$\pm$.002} & 0.704\,{\scriptsize$\pm$.010} & 0.787\,{\scriptsize$\pm$.009} & 0.735\,{\scriptsize$\pm$.004} & 0.576\\
        FLUX-Kontext & 0.780 & 0.554\,{\scriptsize$\pm$.013} & 0.820\,{\scriptsize$\pm$.002} & 0.613\,{\scriptsize$\pm$.011} & 0.786\,{\scriptsize$\pm$.008} & 0.693\,{\scriptsize$\pm$.005} & 0.540\\
        OmniGen2 & 0.832 & 0.565\,{\scriptsize$\pm$.012} & 0.856\,{\scriptsize$\pm$.005} & 0.557\,{\scriptsize$\pm$.015} & 0.775\,{\scriptsize$\pm$.009} & 0.688\,{\scriptsize$\pm$.006} & 0.539\\
        InstructPix2Pix & 0.521 & 0.387\,{\scriptsize$\pm$.014} & 0.770\,{\scriptsize$\pm$.008} & 0.703\,{\scriptsize$\pm$.015} & \textbf{0.881\,{\scriptsize$\pm$.008}} & 0.685\,{\scriptsize$\pm$.006} & 0.344\\
        \midrule
        \rowcolor{gray!18}\multicolumn{8}{l}{\textbf{Track B: mask $+$ text}}\\
        Nano-Banana-2-Lite$^\dagger$ & \textbf{0.894} & \textbf{0.638\,{\scriptsize$\pm$.011}} & 0.953\,{\scriptsize$\pm$.001} & \underline{0.817\,{\scriptsize$\pm$.007}} & 0.861\,{\scriptsize$\pm$.007} & \underline{0.817\,{\scriptsize$\pm$.004}} & \textbf{0.757}\\
        GPT-Image-2$^\dagger$ & 0.833 & 0.605\,{\scriptsize$\pm$.012} & 0.962\,{\scriptsize$\pm$.001} & \textbf{0.856\,{\scriptsize$\pm$.006}} & 0.875\,{\scriptsize$\pm$.007} & \textbf{0.824\,{\scriptsize$\pm$.004}} & \underline{0.724}\\
        Qwen-Image-Edit-inpaint & \underline{0.876} & \underline{0.612\,{\scriptsize$\pm$.011}} & 0.901\,{\scriptsize$\pm$.002} & 0.735\,{\scriptsize$\pm$.011} & 0.830\,{\scriptsize$\pm$.008} & 0.769\,{\scriptsize$\pm$.005} & 0.680\\
        FLUX-Kontext-inpaint & 0.805 & 0.569\,{\scriptsize$\pm$.012} & \underline{0.964\,{\scriptsize$\pm$.001}} & 0.811\,{\scriptsize$\pm$.006} & 0.876\,{\scriptsize$\pm$.007} & 0.805\,{\scriptsize$\pm$.004} & 0.678\\
        BAGEL-7B & 0.785 & 0.538\,{\scriptsize$\pm$.013} & 0.856\,{\scriptsize$\pm$.006} & 0.731\,{\scriptsize$\pm$.010} & 0.741\,{\scriptsize$\pm$.010} & 0.716\,{\scriptsize$\pm$.006} & 0.570\\
        SD3.5-inpaint & 0.841 & 0.559\,{\scriptsize$\pm$.011} & 0.941\,{\scriptsize$\pm$.001} & 0.480\,{\scriptsize$\pm$.013} & 0.733\,{\scriptsize$\pm$.009} & 0.678\,{\scriptsize$\pm$.005} & 0.528\\
        OmniGen2-inpaint & 0.733 & 0.506\,{\scriptsize$\pm$.013} & 0.819\,{\scriptsize$\pm$.006} & 0.507\,{\scriptsize$\pm$.014} & 0.786\,{\scriptsize$\pm$.009} & 0.654\,{\scriptsize$\pm$.006} & 0.449\\
        FLUX.1-Fill & 0.551 & 0.407\,{\scriptsize$\pm$.014} & \textbf{0.974\,{\scriptsize$\pm$.001}} & 0.764\,{\scriptsize$\pm$.009} & \underline{0.881\,{\scriptsize$\pm$.007}} & 0.756\,{\scriptsize$\pm$.005} & 0.446\\
        SDXL-inpaint & 0.477 & 0.350\,{\scriptsize$\pm$.014} & 0.941\,{\scriptsize$\pm$.001} & 0.568\,{\scriptsize$\pm$.013} & \textbf{0.904\,{\scriptsize$\pm$.006}} & 0.691\,{\scriptsize$\pm$.005} & 0.330\\
        \midrule
        \rowcolor{gray!18}\multicolumn{8}{l}{\textbf{Track C: mask $+$ reference}}\\
        Nano-Banana-2-Lite$^\dagger$ & \textbf{0.871} & \textbf{0.642\,{\scriptsize$\pm$.012}} & 0.945\,{\scriptsize$\pm$.002} & \underline{0.808\,{\scriptsize$\pm$.008}} & 0.834\,{\scriptsize$\pm$.008} & \underline{0.807\,{\scriptsize$\pm$.004}} & \textbf{0.726}\\
        GPT-Image-2$^\dagger$ & \underline{0.795} & \underline{0.609\,{\scriptsize$\pm$.013}} & \underline{0.948\,{\scriptsize$\pm$.002}} & \textbf{0.817\,{\scriptsize$\pm$.008}} & \textbf{0.857\,{\scriptsize$\pm$.008}} & \textbf{0.808\,{\scriptsize$\pm$.005}} & \underline{0.675}\\
        FLUX-Kontext-ref & 0.637 & 0.538\,{\scriptsize$\pm$.017} & \textbf{0.959\,{\scriptsize$\pm$.001}} & 0.440\,{\scriptsize$\pm$.013} & 0.718\,{\scriptsize$\pm$.010} & 0.664\,{\scriptsize$\pm$.006} & 0.386\\
        Paint-by-Example & 0.453 & 0.349\,{\scriptsize$\pm$.014} & 0.921\,{\scriptsize$\pm$.001} & 0.600\,{\scriptsize$\pm$.013} & \underline{0.849\,{\scriptsize$\pm$.008}} & 0.680\,{\scriptsize$\pm$.006} & 0.322\\
        Qwen-Image-Edit-ref & 0.510 & 0.423\,{\scriptsize$\pm$.017} & 0.702\,{\scriptsize$\pm$.012} & 0.527\,{\scriptsize$\pm$.018} & 0.827\,{\scriptsize$\pm$.010} & 0.620\,{\scriptsize$\pm$.009} & 0.307\\
        OmniGen2-ref & 0.353 & 0.318\,{\scriptsize$\pm$.016} & 0.739\,{\scriptsize$\pm$.006} & 0.518\,{\scriptsize$\pm$.015} & 0.843\,{\scriptsize$\pm$.009} & 0.605\,{\scriptsize$\pm$.006} & 0.204$^{*}$\\
        BAGEL-7B & 0.612 & 0.504\,{\scriptsize$\pm$.016} & 0.202\,{\scriptsize$\pm$.004} & 0.044\,{\scriptsize$\pm$.004} & 0.525\,{\scriptsize$\pm$.010} & 0.319\,{\scriptsize$\pm$.005} & 0.062$^{*}$\\
        \bottomrule
    \end{tabular*}
    \caption{
        Per-track leaderboard on bench-v1.0 ($1{,}421$ tasks, higher is better), ordered by our primary aggregate \emph{gate} (Sec.~\ref{sec:protocol}); \emph{raw} is the unweighted mean, ``family acc.'' strict SigLIP2 family accuracy, and $\pm$ a 95\% bootstrap interval.
        \textbf{Bold}/\underline{underline}: best/second per column within a track; $^\dagger$ closed; $^{*}$ below the gated do-nothing floor ($0.236$).
        Closed editors lead all three tracks; on Track~C every family loses ground and two entries fall below that floor.
    }
    \label{tab:board}
\end{table*}

\paragraph{Leaderboard.}
Table~\ref{tab:board} reports per-track results. On Track~A the two closed editors lead on both aggregates and clear the exemplar anchor under the gate ($0.760$/$0.739$ vs.\ $0.729$; $0.817$/$0.827$ vs.\ $0.829$ raw). GPT-Image-2 generated the data, so its match to its own exemplars is self-consistency; the informative reading is that Nano-Banana-2-Lite, from a different provider, is the one that ranks first. Closed instruction editors thus reach the data-generating process \emph{without any spatial guidance}, with any home-field advantage bounded by the cross-provider gap. The open field is cleanly separated by architecture: the 20B multi-image editor (Qwen, $0.677$) leads, the unified models BAGEL ($0.576$) and OmniGen2 ($0.539$) bracket the flow-matching editor (FLUX-Kontext, $0.540$), and the SD1.5-era baseline (InstructPix2Pix, $0.344$) trails. Track~B preserves the closed-model order and promotes mask-specialized pipelines (Qwen and FLUX-Kontext inpainting, $0.680$/$0.678$). Track~C inverts the picture: every family drops, and two entries, OmniGen2-ref ($0.204$) and BAGEL-7B ($0.062$), fall below the do-nothing floor ($0.236$).

\section{Analysis}
\label{sec:analysis}

\paragraph{What is a mask worth? (A$\to$B, same model).}
Because tracks share checkpoints, Table~\ref{tab:cross} reads off the causal value of conditioning ($p$ Holm-corrected Wilcoxon throughout). The mask's value spans $0.15$ across the six aligned families: $+0.111$ for FLUX-Kontext, whose Track-A weakness is precisely preservation ($0.82$) and whose inpainting pipeline enforces it ($0.96$, structure riding along $0.61{\to}0.81$); $+0.009$ for Qwen ($\ddagger$); statistically \emph{zero} for both closed editors ($p_{\mathrm{Holm}}{=}0.45$), which already preserve from text alone; and \emph{negative} for both unified models ($-0.019$ BAGEL $\ddagger$, $-0.034$ OmniGen2), which fail to honor it. The gate replicates and sharpens this pattern: FLUX grows to $+0.137$ and OmniGen2 falls to $-0.090$, while BAGEL's deficit loses significance. Masks help only models both weak at preservation \emph{and} able to consume them.

\begin{figure}[t]
    \centering
    \includegraphics[width=\columnwidth]{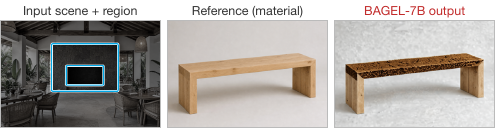}
    \caption{
        Role-binding failure: the model repaints the \emph{reference} (the oak bench), composites the mask silhouette into it, and discards the scene.
    }
    \label{fig:bagelfail}
\end{figure}

\paragraph{What does a reference cost? (B$\to$C, same model).}
Substituting the reference for the instruction degrades \emph{every} family under both aggregates (Table~\ref{tab:cross}). 
\textbf{(1)~Appearance-grounding loss} (closed editors, $-0.010$\,--\,$-0.017$, $\ddagger$-flagged): they still solve the task, just less precisely when the material is shown rather than named. 
\textbf{(2)~Material-identification collapse} (Qwen $-0.150$, OmniGen2 $-0.050$): with no material named in text, family accuracy falls $0.876{\to}0.510$ (Qwen) and $0.733{\to}0.353$ (OmniGen2). These models cannot translate reference \emph{pixels} into the semantic target their editing pathway needs, though the pathway executes correctly when the target arrives as text. Nor is this a classifier failure: the same classifier moves the closed editors only $0.024$--$0.038$ from B to C, and the identity anchor's $0.254$ family score is a task prior (originals often already match the target family), not an error rate. 
\textbf{(3)~Role-binding failure} (BAGEL, $0.319$): the model misassigns the roles of the three input images and repaints the \emph{reference} instead of the scene (Fig.~\ref{fig:bagelfail}), collapsing preservation ($0.202$) and structure ($0.044$). FLUX-Kontext-ref sits between (2) and (3): it keeps the scene ($\mathrm{pres}=0.959$) but pastes unshaded texture, driving structure to $0.440$, near the paste anchor. The prompts name each image's role explicitly and the other five families never swap roles under the identical template, so the collapse belongs to the model. 
The open problem is thus not \emph{rendering} a material, which Track~A shows models do from text, but \emph{grounding a specification given as pixels}.

\paragraph{Practical guidance.}
Where instructions can name the material, closed editors already operate at the exemplar level and masks add friction without accuracy. Where only open models are deployable, the best open entry is a mask-specialized pipeline (FLUX-Kontext-inpaint, $0.805$). Reference-driven specification should not ship today: recognize the swatch to a \emph{name} and route it through text.

\paragraph{Overall ties hide opposite failure modes.}
OmniGen2 and InstructPix2Pix tie on Track A ($0.688$ vs.\ $0.685$, $p_{\mathrm{Holm}}{=}0.86$) for opposite reasons: OmniGen2 gets the material (family $0.83$) but breaks geometry ($\sstr\,0.56$), InstructPix2Pix preserves geometry ($0.70$) but rarely produces the material (family $0.52$). SDXL/SD3.5 on Track~B repeat the pattern ($0.691$ vs.\ $0.678$; $|\Delta|<0.02$, single-seed): SDXL blends beautifully (harmony $0.904$) but paints the wrong material ($0.477$), SD3.5 names it ($0.841$) but deforms the surface (structure $0.480$). A holistic score erases these distinctions; the decomposition makes the board actionable.

\paragraph{Closed vs.\ Open.}
The closed--open gap is concentrated in two dimensions: preservation ($0.95$--$0.96$ vs.\ $0.77$--$0.90$ on Track~A) and structure ($0.82$--$0.85$ vs.\ $0.56$--$0.71$). Material is \emph{not} the differentiator ($0.62$--$0.64$ vs.\ up to $0.62$): open models know \emph{what} to paint but damage the scene while painting it. The gap is engineering, not material knowledge.

\paragraph{Where edits are hard (stratified, Track A).}
Taxonomy labels let us read the board by stratum. Difficulty tracks \emph{mask coverage}: on the $188$ large masks every model drops $0.05$--$0.10$ (GPT-Image-2 $0.755$ $[0.742, 0.768]$ vs.\ $0.840$ $[0.835, 0.845]$ on the $784$ medium; the intervals are disjoint), a bigger region giving geometry and lighting more room to drift. Cluttered ``other'' surfaces are hardest (GPT $0.698$), flat countertops easiest ($0.868$); leather and rattan are the hardest material classes (both $\approx0.77$ for GPT), wood the easiest ($0.881$). Order is stable across strata, except BAGEL-7B is nearly \emph{flat} across materials ($0.72$--$0.75$, overlapping intervals) where others swing $0.10$--$0.15$. The full per-stratum grid ships with the release.

\begin{table}[t]
    \centering
    \scriptsize
    \setlength{\tabcolsep}{3pt}
    \renewcommand{\arraystretch}{0.93}
    \begin{tabular*}{\linewidth}{@{\extracolsep{\fill}}lcccc}
    \toprule
     & \multicolumn{2}{c}{$\Delta_{\mathrm{mask}}$ (A$\to$B)} & \multicolumn{2}{c}{$\Delta_{\mathrm{ref}}$ (B$\to$C)}\\
    \cmidrule(lr){2-3}\cmidrule(lr){4-5}
    family & raw & gate & raw & gate\\
    \midrule
    GPT-Image-2$^\dagger$        & $-0.003^{\mathrm{ns}}$ & $-0.016^{\mathrm{ns}}$ & $-0.017^{\ddagger}$ & $-0.048$\\
    Nano-Banana-2-Lite$^\dagger$ & $0.000^{\mathrm{ns}}$  & $-0.003^{\mathrm{ns}}$ & $-0.010^{\ddagger}$ & $-0.031$\\
    Qwen-Image-Edit              & $+0.009^{\ddagger}$ & $+0.003^{\ddagger}$ & $-0.150$ & $-0.373$\\
    BAGEL-7B                     & $-0.019^{\ddagger}$ & $-0.006^{\mathrm{ns}}$ & $-0.397$ & $-0.508$\\
    FLUX-Kontext                 & $+0.111$ & $+0.137$ & $-0.141$ & $-0.291$\\
    OmniGen2                     & $-0.034$ & $-0.090$ & $-0.050$ & $-0.245$\\
    \bottomrule
    \end{tabular*}
    \caption{Conditioning-aligned cross-track deltas: adding a mask ($\Delta_{\mathrm{mask}}$) and replacing text with the reference ($\Delta_{\mathrm{ref}}$), as task-paired means under both aggregates (per-track scores in Table~\ref{tab:board}). Unmarked entries are Holm-corrected Wilcoxon $p<10^{-3}$; $^{\mathrm{ns}}$ not significant; $^{\ddagger}$ marks $|\Delta|<0.02$ from a single seed, significant over tasks but not seeds (Sec.~\ref{sec:protocol}). Beyond $\ddagger$, the mask helps one family and hurts one; the reference hurts \emph{all six}, more under the gate.}
    \label{tab:cross}
\end{table}

\section{Human Expert Calibration}
\label{sec:human}
As a calibration pilot, two expert raters produced 459 two-alternative
forced-choice judgments over 422 unique tasks, with hidden
identities, randomized sides, optional ties, and per-dimension failure
tags. The 34 anchor--anchor pairings are held out of scoring; the 25 with
winners known by construction serve as attention checks (24 passed, one
tied). We fit Bradley--Terry strengths to the remaining 425 scoring
judgments (ties as half-wins, bootstrap $B{=}2{,}000$).

\paragraph{Ranking agreement.}
Across the 20 entries with at least ten judgments, human strengths correlate
with the leaderboard at Kendall $\tau=0.68$ and Spearman $\rho=0.87$. On
Track A, $\tau$ rises from $0.62$ to $0.87$ once the single outlier
discussed below is excluded ($1.00$ under the gate). Against the exemplar, GPT-Image-2 ties (5W--4L--12T for the model) and
Nano-Banana-2-Lite is preferred outright (8W--4L--7T); in the direct
closed-vs-closed pairing the raters prefer the \emph{non-coupled} editor
(5--2--10), so generator identity buys no human-visible advantage. The
Track-C degradation replicates family by family with the metric's severity
ordering: near-unanimous for the two largest drops (Qwen 18--1,
FLUX-Kontext 16--1--2), mild for the closed pair (3--1--4, 3--0--7), with
one reversal (OmniGen2, 2--4--3) resting on nine judgments. Metric
near-ties are human near-ties: GPT-Image-2 Track A vs.\ B
($\Delta_{\mathrm{mask}}$ n.s.) draws seven ties in nine pairings, and
OmniGen2 vs.\ InstructPix2Pix (gap $0.003$) splits 12--7 (n.s.). The 28\%
tie rate falls on the pairings the leaderboard also refuses to separate.

\begin{table}[t]
    \centering
    \scriptsize
    \setlength{\tabcolsep}{3pt}
    \renewcommand{\arraystretch}{0.93}
    \begin{tabular*}{\linewidth}{@{\extracolsep{\fill}}lccccccc}
    \toprule
     & \multicolumn{3}{c}{Ours} & \multicolumn{4}{c}{Established}\\
    \cmidrule(lr){2-4}\cmidrule(lr){5-8}
     & gate & raw & fam. & $-$LPIPS & CLIP-I & CLIPScore & CLIP$_{\mathrm{dir}}$\\
    \midrule
    $\tau_{\mathrm{H}}$ & \textbf{0.72} & 0.68 & 0.64 & 0.55 & 0.54 & 0.05 & $-0.44$\\
    $\rho_{\mathrm{H}}$ & \textbf{0.89} & 0.87 & 0.86 & 0.75 & 0.76 & 0.15 & $-0.52$\\
    \bottomrule
    \end{tabular*}
    \caption{Kendall $\tau_{\mathrm{H}}$ and Spearman $\rho_{\mathrm{H}}$ against human Bradley--Terry strengths (425 judgments); GT-referenced baselines use the 1{,}093 clean-exemplar tasks. Ours exceeds every established metric: significantly for LPIPS and CLIPScore (bootstrap $P\geq0.985$), marginally for CLIP-I ($\Delta\tau=+0.14$, CI $[-0.01,0.28]$).}
    \label{tab:baselines}
\end{table}

\paragraph{Against existing metrics.}
The same judgments adjudicate between our protocol and the metrics it
critiques (Table~\ref{tab:baselines}): every aggregate of ours correlates
with the raters more strongly than every baseline, significantly so over
LPIPS-to-exemplar and CLIPScore, marginally over CLIP-I. CLIPScore spans
only $0.019$ across the 20 systems: text alignment cannot separate a
correct swap from repainting the room.
Directional CLIP is \emph{anti}-correlated: it rewards large moves toward
the target material, and the systems humans rank last make exactly such
moves while destroying the scene. The GT-referenced pair scores the
exemplar perfectly by construction, where the raters place it fifth of 20.

\paragraph{Divergences.}
The disagreements are as informative as the agreements and share one
theme: raters weight the edited region (material identity, shading
realism) and discount the outside-region fidelity two dimensions reward. The metric places BAGEL-7B
fourth on Track A; raters rank it last (1--14 vs.\ Qwen and FLUX-Kontext), its
losses drawing the most lighting tags, the sole driver of the Track-A $\tau$
gap. FLUX-Kontext's $+0.111$ mask gain is invisible (8--1--13): the mask
repairs measured fidelity, not perceived quality. Raters prefer paste to doing
nothing (6--3) and Qwen-ref above FLUX-ref (7--4--1), punishing sticker pastes
harder than misidentification.
Per-pair $n$ is 9--20; the OmniGen2 reversal and conservatism probe rest on nine.

\section{Limitations}
The benchmark is fully synthetic and single-source, disclosed: all four task images come from GPT-Image-2, also evaluated. The reference-free protocol removes source bias from \emph{scoring} and the cross-provider gap ($\leq0.010$) bounds home-field advantage, but admission conditions on the generator drawing an aligned mask (a selection bias), and transfer to real photographs is unverified. Human agreement is a first answer (Sec.~\ref{sec:human}) from two raters with non-overlapping judgments; anchor checks substitute for an inter-rater statistic. Mask quality caps yield at $70.8\%$; the curation VLM shares the generator's provider, under-representing tasks illegible to it; and the taxonomy is furniture-heavy (825 of 1{,}421), so wall and countertop conclusions rest on smaller strata. Finally, raw \texttt{overall} rewards conservatism (identity $0.810$), which the gate view mitigates in Table~\ref{tab:robust}.

\section{Reproducibility}
We release everything\footnote{\url{https://anonymous.4open.science/r/MatReplace}}, including task specifications, the benchmark harness, model generations with per-task scores~($\sim$33k images) and the full human study. Closed models are pinned by provider ID and date, open models by checkpoint hash and sampler, libraries by a lockfile.

\section{Conclusion}
MatReplace isolates material replacement, scores it without a ground-truth reference, and aligns model families across three conditioning tracks. Three findings emerge: closed editors match the data-generating process without spatial guidance; the benefit of masks disappears with model strength and can reverse; and reference-driven specification remains open, degrading every family. A two-rater expert study reproduces the ranking and prefers our aggregates to the GT-referenced and CLIP baselines, and the aligned tracks resist gaming by imitation.

\bibliography{references}


\end{document}